\documentclass{article}
\usepackage{spconf,amsmath,graphicx,hyperref}
\usepackage{amssymb}
\usepackage{mathtools}
\usepackage{amsthm}
\usepackage{booktabs}

\title{Align2Speak: Improving TTS for Low Resource Languages via ASR-Guided Online Preference Optimization}
\name{\begin{tabular}{c}
    Shehzeen Hussain, Paarth Neekhara, Xuesong Yang, Edresson Casanova, Subhankar Ghosh \\ 
    Roy Fejgin, Ryan Langman, Mikyas Desta, Leili Tavabi, Jason Li
  \end{tabular}}

\address{NVIDIA Corporation, USA }
\begin{document}
%\ninept
%
\maketitle
\begin{abstract}
Developing high-quality text-to-speech~(TTS) systems for low-resource languages is challenging due to the scarcity of paired text and speech data. In contrast, automatic speech recognition~(ASR) models for such languages are often more accessible, owing to large-scale multilingual pre-training efforts.
We propose a framework based on Group Relative Policy Optimization~(GRPO) to adapt an autoregressive, multilingual TTS model to new languages. 
Our method first establishes a language-agnostic foundation for TTS synthesis by training a multilingual baseline with International Phonetic Alphabet (IPA) tokens. Next, we fine-tune this model on limited paired data of the new languages to capture the target language's prosodic features.
Finally, we apply GRPO to optimize the model using only unpaired text and speaker prompts, guided by a multi-objective reward from pretrained ASR, speaker verification, and audio quality estimation models.
Experiments demonstrate that this pipeline produces intelligible and speaker-consistent speech in low-resource languages, substantially outperforming fine-tuning alone. 
Furthermore, our GRPO-based framework also improves TTS performance in high-resource languages, surpassing offline alignment methods such as Direct Preference Optimization~(DPO) yielding superior intelligibility, speaker similarity, and audio quality.\footnote{Project Webpage: \url{grpotts.github.io }}

\end{abstract}
\vspace{-2mm}
\section{Introduction}
\label{sec:intro}

Recent advances in neural text-to-speech (TTS) synthesis have led to highly natural and intelligible systems, enabled by large-scale datasets and powerful generative models such as autoregressive transformers and diffusion models. Particularly, large language model (LLM) based TTS approaches have become increasingly common, leveraging the strong representation learning capabilities of LLMs to improve prosody, context adherence and naturalness of generated speech~\cite{borsos2023audiolm,t5tts,yanguniaudio,song2025ella,wang2024speechx,koeltts}.
However, the development of TTS for low-resource languages remains a pressing challenge. Collecting high-quality paired text–speech corpora is expensive and often infeasible for many underrepresented languages. 
This data scarcity creates a substantial performance gap compared to high-resource languages~\cite{chen2019end,lux2022low}.

In contrast, automatic speech recognition (ASR) systems for low-resource languages have benefited significantly from large-scale multilingual pretraining and transfer learning. Past work such as Whisper~\cite{radford2023robust} and Common Voice~\cite{ardila2020common} have made robust ASR models widely available, even for languages with limited training resources. 
This asymmetry suggests an opportunity: while curating extensive TTS data is difficult, one may still obtain a reliable ASR model for a given low-resource language, which can serve as a feedback signal for improving TTS models.

\begin{figure*}[htp]
\vspace{-8mm}
    \centering
    \includegraphics[width=0.95\textwidth]{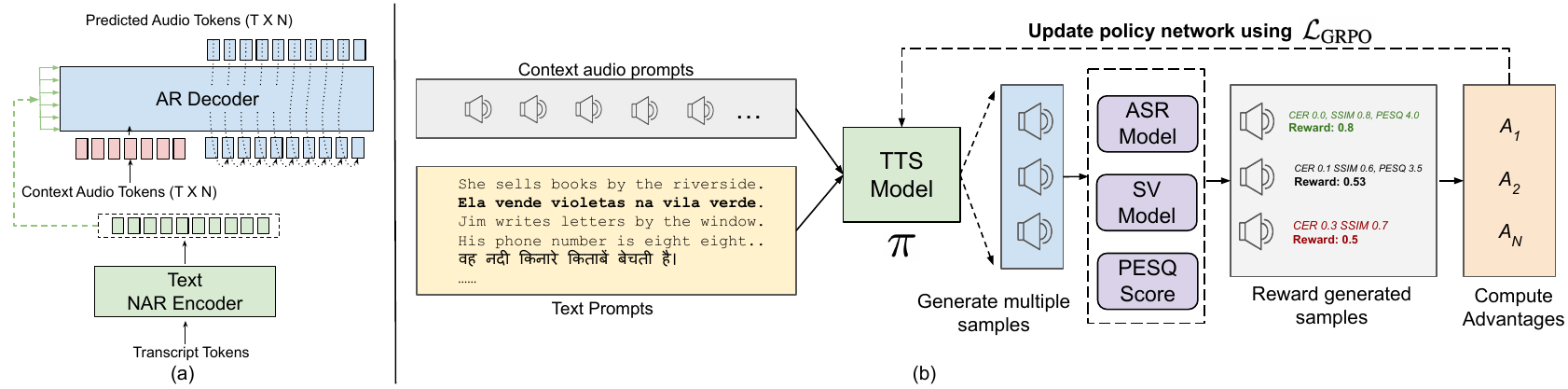}
\vspace{-4mm}    
\caption{\footnotesize{(a) Encoder-Decoder TTS Model Architecture, (b) Online Preference Alignment for TTS: Model generates multiple outputs for a text and context audio prompt, which are then rewarded using ASR, SV models and PESQ Scores and optimized using GRPO.}}
\vspace{-2mm}
\label{figs:grpo_diag}
\end{figure*}

A parallel line of research in natural language processing has shown that reinforcement learning (RL) can effectively align generative models with task-specific objectives. Reinforcement learning from human feedback (RLHF) has emerged as a central paradigm for large language models (LLMs), where an external ``judge” (human or automated) provides preference signals that guide the model towards producing more useful outputs. Recently, similar ideas have been explored in speech generation.
Offline preference optimization methods such as Direct Preference Optimization (DPO)~\cite{rafailov2023direct} have been applied to TTS synthesis~\cite{koeltts,zhang2024speechalign}, but these approaches lack the dynamic feedback loop provided by online RL methods.

In this work, we investigate whether \textit{ASR models can be leveraged as a reward function for learning TTS in low-resource settings}. 
Specifically, we introduce a framework based on \textit{Group Relative Policy Optimization (GRPO)}~\cite{shao2024deepseekmath}, where an autoregressive TTS model trained across multiple languages is adapted to new languages using reinforcement learning. Our method proceeds in three stages:  1) We train a baseline autoregressive TTS model on a set of source languages, using the International Phonetic Alphabet (IPA) as the text representation. 2) The baseline model is adapted to a new target language using a small amount of available paired data, capturing its phonetic and prosodic patterns. 3) We generate text and speaker prompts in the target language and optimize the TTS model across multiple objectives using GRPO, guided by reward metrics from ASR, speaker verification, and PESQ~\cite{kumar2023torchaudio} estimation models to improve intelligibility, speaker consistency, and perceptual quality.

Our experiments show that this pipeline of fine-tuning followed by GRPO with automatic reward signals, yields intelligible and speaker-consistent TTS for new low-resource languages, significantly outperforming fine-tuning alone. Furthermore, we show that GRPO provides strong preference alignment even in \textit{high-resource (seen) languages}, outperforming offline alignment methods such as DPO. 
To the best of our knowledge, this is the first work to extend GRPO to token-based TTS synthesis with multi-objective rewards.
% , showing that online reinforcement learning provides superior preference alignment compared to prior approaches that rely on offline alignment methods such as DPO. 
These findings suggest that online reinforcement learning offers a general and effective mechanism for aligning TTS models with perceptual and linguistic objectives.

% In summary, the contributions of this paper are as follows:  

\vspace{-2mm}
\section{Methodology}
\label{sec:methodology}
This section outlines our proposed framework for adapting a multilingual text-to-speech (TTS) model to a low-resource language using the three-step framework described above.
% . Our approach consists of three main stages: (1) pretraining a multilingual baseline model using a phonetic input representation for cross-lingual generalization, (2) fine-tuning the model on limited paired data from the target language, and (3) further refining the model using GRPO, with rewards derived from automatic speech recognition (ASR), speaker verification and PESQ estimator models.

\vspace{-4mm}
\subsection{Baseline TTS Model}
Our synthesis framework is based on Koel-TTS~\cite{koeltts}, an autoregressive encoder-decoder TTS model that operates on the low-frame rate ($21.5$ FPS) audio codec representation given by NanoCodec~\cite{casanova2025nanocodec}.  
Koel-TTS comprises an autoregressive (AR) transformer decoder conditioned on text encodings from a non-autoregressive (NAR) transformer encoder using cross-attention (Figure~\ref{figs:grpo_diag}a).
Specifically, given an input text sequence and acoustic codes of a reference speaker audio, the TTS model autoregressively generates the acoustic codes corresponding to the input text and speaker. 
To build a foundation model capable of generalizing to new languages, we first pretrain the Koel-TTS model on a multilingual dataset comprising six languages --- English, Dutch, Italian, Spanish, French, and German. 
A key aspect of our pretraining strategy is the use of the International Phonetic Alphabet (IPA) for text representation~\cite{international1999handbook}. 
IPA provides a universal, standardized representation of speech sounds. 
The IPA text is tokenized using a byte-level tokenizer, which naturally handles the full range of IPA symbols with just $256$ tokens. 
This phonetic-based approach enables the model to learn a mapping from universal acoustic units to speech, which is crucial for zero-shot or few-shot adaptation to a new language. 
The baseline model is trained on triplets of \emph{context audio, IPA transcript, and target audio} and optimized using only the next-frame prediction loss with a parallel head~\cite{t5tts,koeltts}. 
% Unlike past work~\cite{} we do not use any auxiliary CTC loss on the attention heads, since IPA 

\vspace{-4mm}
\subsection{Fine-Tuning on Low-Resource Languages}
After pretraining, we adapt the baseline model to a target low-resource language through supervised fine-tuning. This stage uses a limited amount of paired data (i.e., a few hours of speech and corresponding transcripts) from the target language. 
To prevent catastrophic forgetting of the knowledge acquired during pretraining, we employ a mixed-data training strategy. 
The fine-tuning dataset is a combination of the original pretraining data and the new low-resource data. 
We upsample the low-resource data, ensuring that each training batch contains a high proportion of samples from the target language. 
This approach allows the model to learn the specific phonetic and prosodic characteristics of the new language while retaining its general synthesis capabilities for languages seen during baseline model training. 
The resulting fine-tuned model serves as the reference policy for the subsequent reinforcement learning stage.

\vspace{-4mm}
\subsection{Refinement with GRPO}
\label{sec:rewardsetup}

While fine-tuning adapts the model to the target language, its quality is often limited by the small data size. To further enhance intelligibility and speaker similarity, we adapt Group Relative Policy Optimization (GRPO)~\cite{shao2024deepseekmath}, an online reinforcement learning algorithm designed for policy alignment. This procedure is depicted in Figure~\ref{figs:grpo_diag}(b).

\noindent \textbf{Prompt Construction:}
The GRPO training process requires prompts from which the model generates speech. 
Each prompt consists of a speaker reference audio and a text transcript to be synthesized. 
For low-resource scenarios, a significant advantage of this setup is that the content of the speaker audio prompt does not need to match the text transcript. 
This decouples the need for paired data, allowing us to leverage readily available, non-parallel speech and text data for creating a set of prompts. Specifically, we include $15k$ text and speaker reference audio pairs as the prompts for each language in our dataset. During GRPO, we construct prompts for all languages, including those seen during baseline model training. 

\noindent \textbf{Reward Function:}
For each generated audio sample, we compute a reward signal $R$ that quantifies its preference. 
The total reward is a weighted combination of three components: transcription accuracy, measured by Character Error Rate~(CER), speaker similarity, measured by a cosine similarity score~(SSIM) and perceptual speech quality, measured by a neural PESQ estimator:
\begin{equation}
    R = w_\textit{cer} \cdot R_\textit{cer} + w_\textit{ssim} \cdot R_\textit{ssim} + w_\textit{pesq} \cdot R_\textit{pesq}
\end{equation}
where we set $w_\textit{cer} = w_\textit{ssim} = 0.45$ and $w_\textit{pesq}=0.1$.

\begin{figure}[!ht]
    \centering
    \includegraphics[width=1.0\columnwidth]{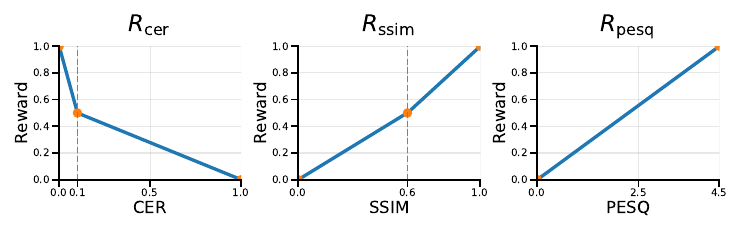}
    \vspace{-8mm} 
    \caption{\footnotesize{Reward functions used to map CER, SSIM and PESQ values to normalized rewards between $0$ and $1$.}}
    \label{figs:rewardgraphs}
     \vspace{-2mm} 
\end{figure}

To ensure the reward signal is well-scaled and interpretable, we map the raw CER and SSIM scores to a normalized reward value between $0$ and $1$ using a piecewise linear function. This function is defined by three key points: the worst possible score maps to 0, the best possible score maps to 1, and the average score observed from the baseline model maps to 0.5. The reward for intermediate values is linearly interpolated. 
To optimize for PESQ, we normalize the raw PESQ score between $0$ and $1$ by setting $R_\textit{pesq}=\textit{PESQ}/4.5$. 
% Figure~\ref{figs:rewardgraphs} shows the mapping of the raw CER, SSIM, and PESQ values to their corresponding rewards. 
The maximum CER and minimum PESQ values are clipped at $1$ and $0$ respectively before normalization.
% This scaling stabilizes training by preventing extreme reward values and provides a more consistent gradient signal. 
For measuring CER, we use the multilingual Whisper Large V3 model~\cite{radford2023robust}. SSIM is computed between the context audio and the generated audio using the Titanet-large model~\cite{koluguri2022titanet}. PESQ is estimated using the reference-free neural PESQ estimator from torchaudio~\cite{kumar2023torchaudio}. The precise reward normalization functions used in our framework are shown in Figure~\ref{figs:rewardgraphs}.

\vspace{-4mm}

\noindent \textbf{GRPO Objective:}
GRPO optimizes the policy $\pi_{\theta}$ (our TTS model) to maximize the expected reward from generated samples~\cite{shao2024deepseekmath}. For a given text and context audio input $x=(x_\textit{text}, x_\textit{audio})$, the model's response distribution $\pi(y|x)$ encompasses a range of potential outputs $y$ with varying reward values. 
% GRPO objective maximizes the expected reward as follows:
% \begin{equation}
% \label{eq:grpo_objective}
% J(\pi_\theta) = \mathbb{E}_{x \sim \mathcal{D}, y \sim \pi_\theta(\cdot|x)} [R(x, y)] - \beta D_{KL}(\pi_\theta(\cdot|x) || \pi_{ref}(\cdot|x))
% \end{equation}
% Here, $x$ is the prompt sampled from the distribution $\mathcal{D}$, $y$ is the audio generated by the current policy $\pi_{\theta}$, $R(x,y)$ is our scaled reward function, and $\beta$ is a hyperparameter that controls the strength of the KL penalty. This objective encourages the model to explore generations that yield higher rewards (i.e., more intelligible and speaker-consistent speech) without sacrificing the overall quality and stability learned during the pretraining and SFT stages.
% We fine-tune the policy $\pi_\theta$ using \emph{Group Relative Policy Optimization} (GRPO), a preference-based objective that reduces variance by comparing responses within the same prompt group.  
For each prompt $x_i$, we sample $K$ responses $\{y_{i,k}\}_{k=1}^K \sim \pi_{\theta_{\text{old}}}$ and obtain rewards $r_{i,k}$.  
A group baseline is defined as
\begin{align}
\mu_i &= \frac{1}{K}\sum_{k=1}^K r_{i,k}, \qquad
A_{i,k} \triangleq r_{i,k} - \mu_i.
\end{align}
The training objective of GRPO directly maximizes the log-likelihood of responses weighted by these centered advantages $A_{i,k}$:
\begin{align}
\mathcal{L}_{\text{GRPO}}(\theta)
= \frac{1}{MK}\sum_{i=1}^M \sum_{k=1}^K
A_{i,k}\,\log \pi_\theta(y_{i,k}\mid x_i).
\end{align}

We omit the KL penalty to the reference model proposed in the original GRPO formulation~\cite{shao2024deepseekmath} since we find that relying solely on the group-relative advantages stabilizes learning and speeds up training.

\begin{figure*}[htp]
\vspace{-8mm}
    \centering
    \includegraphics[width=1.0\textwidth]{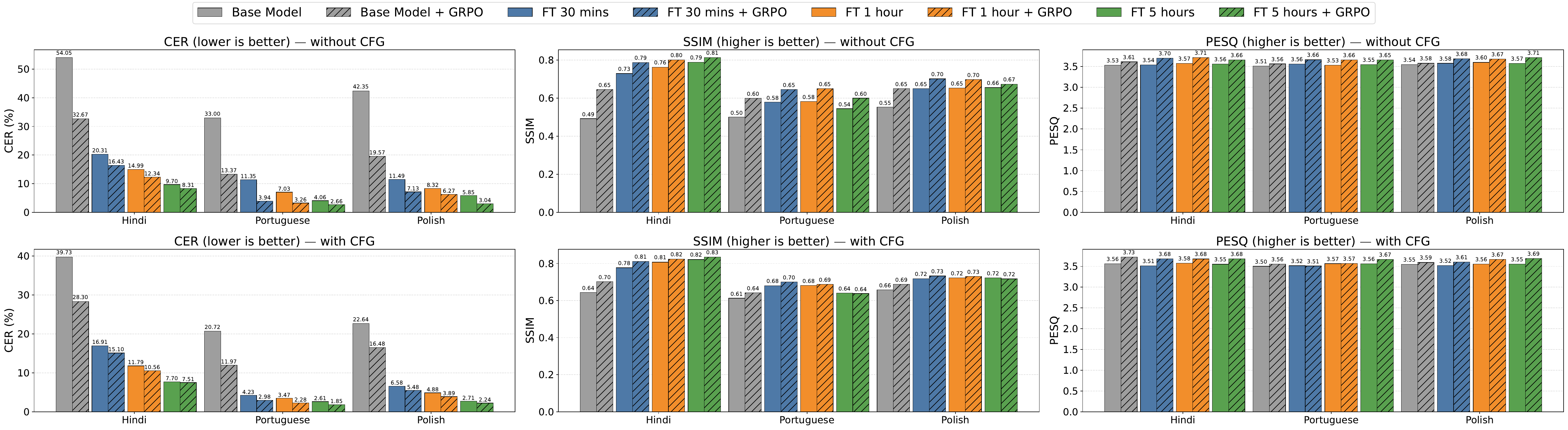}
\vspace{-8mm}    
\caption{\footnotesize{Intelligibility (CER), Speaker similarity (SSIM) and perceptual quality (PESQ) evaluation of the baseline model and the low-resource adaptations for new languages. Post GRPO, we observe considerable improvements in CER across both baseline and fine-tuned models.}}
\vspace{-4mm}
\label{figs:results}
\end{figure*}

\vspace{-2mm}
\section{Experiments}

\subsection{Experiment Setup}
\textbf{Baseline Model Training:} Our baseline TTS model is trained on English and $5$ European languages. For English, the training corpus consists of approximately $18k$ hours of TTS data from the following datasets: \emph{train-clean-360} and \emph{train-clean-100} subsets of LibriTTS~\cite{zen2019libritts}, HiFiTTS~\cite{bakhturina21_interspeech}, a $17k$-hour subset of the LibriVox MLS dataset~\cite{pratap20_interspeech} and a proprietary, 2-speaker, 63-hour dataset. For European languages, we use $5$ languages from the CML dataset~\cite{oliveira2023cml} that contains $1{,}562$ hours of German, $642$ hours of Dutch, $476$ hours of Spanish, $283$ hours of French, $131$ hours of Italian speech data. 
To enable Classifier Free Guidance (CFG) during inference, we follow the methodology proposed in Koel-TTS~\cite{koeltts} to randomly drop out conditioning input during training.

The baseline model is trained on $32$ NVIDIA H100 GPUs using a global batch size of $512$, optimized using Adam optimizer with an initial learning rate of \texttt{1e-4}. The learning rate is annealed every $1k$ training steps using an exponential decay factor of $0.998$. Training for the model converges in around $300k$ iterations.

\noindent \textbf{Finetuning for low-resource languages:} We consider three low-resource languages for our experiments: Polish and Portuguese from the CML dataset~\cite{oliveira2023cml}; and Hindi from the \mbox{IndicTTS} dataset~\cite{baby2016resources}. For our experiments, we create random subsets comprising $30$ minutes, $1$ hour and $5$ hours of data from each of the three low-resource languages for the fine-tuning stage. 
For fine-tuning, we mix the data from the new languages with the $21k$ hours of multilingual data described above. 
To ensure sufficient exposure to the new languages during fine-tuning, we upsample their data by a factor of $5$ compared to their natural proportion in the multilingual corpus. Finetuning is performed at a fixed learning rate of \texttt{1e-5} for a maximum of $30k$ steps using a global batch size of $512$. We choose the checkpoint with the lowest validation loss on a validation set containing only the new languages.

\noindent \textbf{Preference Alignment using GRPO:} 
% For preference alignment, we create a prompt dataset comprising $15k$ text and context audio pairs per language for all languages (baseline model and new languages) in our dataset. 
GRPO is performed for a maximum of $2k$ mini-batch iterations where each mini-batch contains $64$ text and context audio prompts, at a fixed learning rate of \texttt{2e-7}. For each prompt, we generate $12$ audios using multinomial sampling at temperature $0.7$. 
During GRPO, CFG is used for inference with a probability of $0.5$ and CFG scale of $2.5$, to improve model alignment both with and without CFG. 
% The generations are rewarded using the setup described in Section~\ref{sec:rewardsetup}. 
We validate the model every $50$ iterations on a validation set containing the new languages, and choose the checkpoint with the best $R_\textit{CER}$ on this validation set.

\noindent \textbf{Evaluation:}
% The generated speech is evaluated for 
We evaluate the generated speech along three primary dimensions: intelligibility, speaker similarity, and audio quality. 
Intelligibility is quantified using ASR-derived character error rate (CER) using whisper-large-v3~\cite{radford2023robust}. Speaker similarity is measured by computing the cosine similarity (SSIM) between embeddings of synthesized utterances and their corresponding context audio. Embeddings are obtained with Titanet-Small~\cite{koluguri2022titanet}, distinct from the Titanet-Large model applied during preference alignment. Audio quality is measured using a reference-free neural PESQ estimator~\cite{kumar2023torchaudio} and naturalness is evaluated using Squim-MOS~\cite{kumar2023torchaudio}.
% For evaluation on unseen English speakers, we construct a subset from the \textit{test-clean} portion of LibriTTS. This subset comprises $180$ utterances drawn from $36$ of the $40$ available speakers, with $5$ distinct pairs of context and target audios per speaker. During inference, a random 5-second segment from the context audio is used consistently across runs. For non-English evaluations, we select $100$ speaker-balanced utterances per language from the CML test set and IndicTTS test set.
% Inference is carried out using multinomial sampling with a temperature of \texttt{0.6}. Since the method is inherently stochastic, each experimental condition is repeated five times, and we report the mean scores along with 95\% confidence intervals.

% \vspace{-2mm}
% \section{Results}

\vspace{-2mm}
\subsection{Offline vs Online Preference Optimization}
To provide a direct comparison with the DPO approach introduced in Koel-TTS, we apply GRPO to the \textit{Koel-TTS 380M} English TTS model~\cite{koeltts}, using the same prompt dataset. For consistency, we exclude the PESQ reward term, ensuring the setup is directly aligned with the conditions reported for DPO. Our results show that GRPO consistently achieves higher intelligibility and speaker similarity scores while maintaining comparable naturalness. This demonstrates that online preference optimization, which leverages continuous feedback and iterative improvement, is more effective than offline alignment techniques such as DPO.

\vspace{-6mm}
\begin{table}[h]
\centering
\caption{\footnotesize{Comparison of Offline vs Online Preference Optimization for English TTS Synthesis (with and without CFG). Results reported with $95\%$ confidence intervals averaged across $5$ inference runs.}}
\vspace{3mm}
\resizebox{1.0\columnwidth}{!}{%
\begin{tabular}{lcccc}
\toprule
Model & CFG & CER ($\downarrow$) & SSIM ($\uparrow$) & Squim-MOS ($\uparrow$) \\
\midrule
Base & No & $2.68 \pm 1.13$ & $0.637 \pm 0.008$ & $4.35 \pm 0.02$ \\
Base + DPO & No & $0.89 \pm 0.15$ & $0.667 \pm 0.003$ & $4.40 \pm 0.01$ \\
\textbf{Base + GRPO} & No & $\mathbf{0.56 \pm 0.24}$ & $\mathbf{0.759 \pm 0.002}$ & $4.39 \pm 0.01$ \\
\midrule
Base & Yes & $0.57 \pm 0.11$ & $0.720 \pm 0.004$ & $4.41 \pm 0.01 $ \\
Base + DPO & Yes & $0.55 \pm 0.10$ & $0.729 \pm 0.003$ & $4.41 \pm 0.01 $ \\
\textbf{Base + GRPO} & Yes & $\mathbf{0.53 \pm 0.16}$ & $\mathbf{0.783 \pm 0.005}$ & $4.40 \pm 0.01 $ \\
\bottomrule
\end{tabular}
}
\vspace{-4mm}
\label{tab:offline_online_comp}
\end{table}

\subsection{TTS Synthesis for Low Resource Languages}
We present the performance of the baseline, fine-tuned, and preference-aligned models on low-resource languages in Figure~\ref{figs:results}. The baseline model (solid gray), which was not trained on these languages, shows limited generalizability despite IPA tokenization, evident from its high CER scores. Interestingly, applying our multi-objective GRPO directly on this baseline (dashed gray) substantially reduces CER and boosts SSIM, even without access to paired text–speech data in the target language. This suggests that preference alignment alone can bridge part of the gap in learning TTS for new languages.

Fine-tuning on small amounts of target language data further amplifies performance gains. Models trained with just $30$ minutes of supervision already outperform the baseline by a wide margin, with steady improvements as the fine-tuning budget increases to $1$ hour and $5$ hours. Notably, aligning fine-tuned TTS checkpoints with the proposed GRPO objective consistently lowers CER even further, highlighting the complementary benefits of preference alignment.
For example, with only $30$ minutes of Portuguese data, CER drops from $33.00\%$ in the baseline to $3.94\%$ after fine-tuning and GRPO, which is a reduction of more than $8\times$. Similar trends hold across Hindi and Polish, underscoring the robustness of this approach across typologically diverse languages.

Beyond intelligibility, GRPO also enhances SSIM and PESQ scores. GRPO benefits not only unseen languages but also those seen during baseline model training as shown in Table~\ref{tab:offline_online_comp} 
(see project webpage for more details). 
% When 5 hours of data is available, 
Overall, combining small-scale fine-tuning with preference alignment enables strong, scalable adaptation to new languages with minimal data.

\vspace{-6mm}
\section{Conclusion}
In this work, we introduce a GRPO-based framework for adapting multilingual TTS generation models to low-resource languages by leveraging a pretrained ASR model for the new languages. When combining supervised fine-tuning on limited paired data with online preference optimization using multi-objective rewards, our approach delivers consistent gains in intelligibility, speaker similarity, and perceptual quality. Beyond low-resource adaptation, we also demonstrate that GRPO guided by automatic reward functions for TTS, provides stronger preference alignment than offline methods, even for widely used, data-rich languages, highlighting its broad utility for TTS synthesis.

% Overall, these findings show that preference alignment is a powerful complement to low-resource finetuning. Even when limited supervision is available, GRPO enables large gains in intelligibility, similarity, and quality—paving the way for scalable adaptation of speech generation models to new languages. 

\bibliographystyle{IEEEbib}
\bibliography{strings,refs}

\end{document}